\documentclass{llncs}
\usepackage{graphicx}

\begin{document}

\frontmatter

\pagestyle{headings}
\mainmatter
\titlerunning{}

\title{\textbf{Dilatation of Lateral Ventricles with Brain Volumes in Infants with 3D Transfontanelle US}}

\author{Marc-Antoine Boucher \inst{1} \and Sarah Lipp\'e \inst{2,3} \and Am\'elie Damphousse \inst{3} \and Ramy El-Jalbout \inst{3} \and Samuel Kadoury\inst{1,3}}


\institute{MedICAL Laboratory, Polytechnique Montreal, Canada \and NED Laboratory, University of Montreal, Montreal, QC, Canada \and CHU Sainte-Justine Research Center, Montreal, QC, Canada 
\thanks{This work was supported by FRQNT (\#198490), RBIQ (\#5886) and CIHR grants. The authors would like to thank Inga Sophia Knoth and Caroline Dupont from Sainte-Justine Hospital for patient recruitement.}}


\maketitle

\vspace{-0.5cm}
\begin{abstract}%
\small
Ultrasound (US) can be used to assess brain development in newborns, as MRI is challenging due to immobilization issues, and may require sedation.  Dilatation of the lateral ventricles in the brain is a risk factor for poorer neurodevelopment outcomes in infants. Hence, 3D US has the ability to assess the volume of the lateral ventricles similar to clinically standard MRI, but manual segmentation is time consuming. The objective of this study is to develop an approach quantifying the ratio of lateral ventricular dilatation with respect to total brain volume using 3D US, which can assess the severity of macrocephaly.  Automatic segmentation of the lateral ventricles is achieved with a multi-atlas deformable registration approach using locally linear correlation metrics for US-MRI fusion, followed by a refinement step using deformable mesh models. Total brain volume is estimated using a 3D ellipsoid modeling approach. Validation was performed on a cohort of 12 infants, ranging from 2 to 8.5 months old, where 3D US and MRI were used to compare brain volumes and segmented lateral ventricles.  Automatically extracted volumes from 3D US show a high correlation and no statistically significant difference when compared to ground truth measurements. Differences in volume ratios was $6.0 \pm 4.8\%$ compared to MRI, while lateral ventricular segmentation yielded a mean Dice coefficient of $70.8\pm3.6\%$  and a mean absolute distance (MAD) of $0.88\pm0.2$mm, demonstrating the clinical benefit of this tool in paediatric ultrasound.

\end{abstract}%

\section{Introduction}%
For newborns, conditions related to cerebrospinal fluid (CSF) like ventriculomegaly (VM) are common disorders, especially for premature newborns which are frequently associated with VM, white matter injury and intraventricular hemorrhage. For newborns, VM is defined as when atriums of lateral ventricles are greater than 10mm. 
Mild VM is associated with neurodevelopmental disorders (learning disorders, autism and hyperactivity deficit) and arises during fetal brain development which could be detected in ultrasound (US). A previous study demonstrated that prenatal VM for full term newborns could lead to an increase in ventricle, intracranial and cortical grey matter volumes  \cite{gilmore2008prenatal}. Changes in sub-cortical regions of the brain is associated with cognitive development and as such, to include diagnosis accuracy, the clinical assessment for VM should include ventricular-brain ratio.


For infants, non-invasive imaging modalities are required for macrocephaly or premature cases of newborns, as well as cases related to neurosurgery or ischemic incident. Therefore, US is often used in neonates to image the developing brain as it is cost effective and accessible. Recent 3D matrix-array transducers can acquire a volume quasi-instantly and acquisition through the fontanelle may become an alternative to MRI for some volumetric assessments, with previous studies evaluating the lateral ventricles with fairly good reliability using US \cite{gilmore2001infant}. Since manual segmentation is time consuming, an automatic segmentation of the lateral ventricles and brain volume in 3D US can be relevant as an objective measure to assess VM in infants with a safe and  accessible imaging modalities.

A few studies focused on the segmentation of lateral ventricles in neonatal brains with 3D US. Lateral ventricles were segmented semi-automatically in 3D US with an overlap of 78.2\% and mean distances of 0.65mm \cite{qiu2015user}, but require manual initialization with landmarks. The work presented in \cite{qiu2017automatic} showed an automatic approach that successfully segmented the ventricles on newborn cerebral 3D US images (76.7\% Dice), but included patients suffering from intraventricular hemorrhage (IVH) with highly enlarged ventricles. Furthermore, the brain volume, which is essential for ventricular-brain volume ratio computation was not evaluated and there were no statistical comparison performed between 3D US and MRI volumes. This method was also applied to intraventricular hemorrhage cases of newborns, and has yet to be validated on normal and on pathological cases. To our knowledge, no study has been conducted to evaluate total brain volume or ventricular-brain volume ratio automatically in 3D US.

In this paper, we present a novel method to compute the ventricular-brain ratio for the diagnosis of VM in infants from 3D US images. Lateral ventricles are segmented with a combination of multi-atlas and deformable mesh registration approaches, from which the ventricular-brain volume ratio can be computed. Results are compared with ground truth manual segmentations on MRI data, demonstrating the clinical potential in paediatric neuroradiology to quantify ventricular enlargement. The contributions are twofold: (1) a novel optimization scheme based on a dynamic weighting factor in the fusion process, handling hyper and hypo-echoic regions within the ventricles, (2) a geometric-based brain volume estimation method, enabling volume ratios to be extracted, enabling neurodevelopment assessment.

%
\section{Methods}
\label{sec:methodology}

\subsection[1.]{Patient data}
In this study, a cohort of 12 infants aged between 2 and 8.5 months were recruited prospectively, with 3D US and T1 weighted MRI acquired within an hour apart. Ultrasound images were acquired through the fontanelle with an X6-1 matrix-array transducer (EPIQ 7 system, Philips Medical, Bothell, WA) while the MRI was acquired with a 3T MR 750 GE scanner, with a 8 channel head coil, an image resolution of 256 x 256 x 92, and pixel size of 0.78 x 0.78 x 1.2. 3D US was also acquired on 5 additional infants for evaluation purposes. The total brain volume from MRI, which served as ground-truth, was obtained  using the cortical surface extraction sequence of Brainsuite.

\vspace{-0.2cm}
\subsection[2.]{Total brain volume estimation from 3D US}
\label{ssec:subhead}
In cerebral 3D US, the entire brain cannot be fully captured in a single volume even in neonates, due to the size of the transducer and limited acoustic window. Therefore, a total brain volume estimation based on an ellipsoid-fitting method was designed, which doesn't require volume stitching. 

As shown in Fig.1, when fitting a 3D ellipsoid on the skull boundary, the anterior-inferior section of the ellipsoid (shown in dashed lines) overestimates the brain volume. Therefore we estimate the brain volume as a portion of the ellipsoid volume such that:   $V_{brain}=\frac{4}{3} a  b  c  \pi  C_f$ where $a$, $b$ and $c$ are the semi axes of the ellipsoid and $C_f$ is a constant for all patients determined empirically by comparing the ellipsoid and ground truth brain volumes from MRI images. 
To apply the method on 3D US, skull stripping is first applied on the US image as illustrated in Fig.1(c)-(d). The proposed method performs a skull detection based on intensity threshold: $V_{skull}=\{v | I(v) > I_{98}$ and $v \in  A\}$ where $V_{skull}$ is the set of all skull voxels, $v$ a voxel in the 3D US image, $I(v)$ the intensity of this voxel, $I_{98}$ is the 98 percentile of the image intensities and $A$ an area determined from the ellipsoid geometry as follows: 
\begin{equation}\label{ellipsevolume}
A=\{(x,y,z) | 0.8 < \frac{x^2}{a_v} + \frac{y^2}{b_v} + \frac{z^2}{c_v} <1.3\}.
\end{equation}

The centroid position of the brain used to estimate $A$ is constant for all patients based on empiric observations, $z_{center}$ is at 65\% height level of the non-zero intensities, and $x_{center}$ and $y_{center}$ are in the middle of the non-zero intensities in the $z_{center}$ plane. Since the size of the US image is fixed, and the pixel spacing changes according to the brain size, $a_v$, $b_v$ and $c_v$ are the semi axes in fixed voxel size. Finally, the parameters of the ellipsoid shape are optimized to fit the detected boundaries from the overall appearance of the brain's shape. 

Once the shape is obtained, the upper brain limit is approximated with Point 2 as the superior brain limit in 3D US and point 1 is the upper transducer position on the skull which is at the same height as point 2 (Fig. 1).

\begin{figure}[h]
\begin{center}
\includegraphics [scale=0.49]{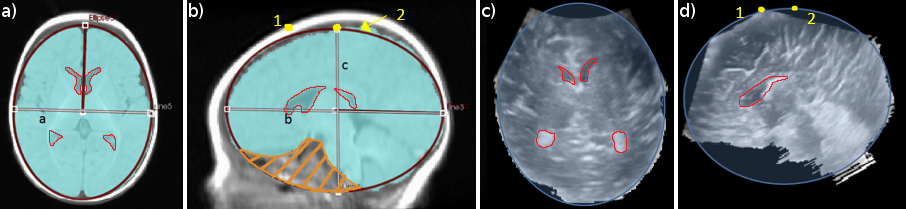}
\caption{\small  (a) Ellipsoid fitting on the MRI axial view; (b) Ellipsoid fitting on the MRI sagittal view; (c) Ellipsoid fitted on the stripped brain in 3D US axial view; (d) Ellipsoid fitted on stripped brain in sagittal view 3D US. Lateral ventricles are outlined in red.}
\end{center}
\vspace{-0.6cm}
\end{figure}
\vspace{-0.1cm}

\subsection[2.]{Lateral ventricular segmentation}
\label{ssec:subhead}
The first step of the lateral ventricular segmentation method is a multi-atlas registration, where MRI atlases are registered to the infant's 3D US image. This is followed by a label fusion where the output is converted to a mesh. Finally, a deformable mesh based segmentation is applied to account for anatomical variabilities not captured by the atlases.
\vspace{-0.3cm}
\subsubsection{Initialization.}
The orientation of the US images is first corrected by rotating the volume to match the orientation observed on the MRI atlases. This is performed from a PCA on the extracted inferior skull region, to identify the principal orientation vectors of the head. Then, the brain in 3D US is extracted with the method described in 2.2 and its center position and size are calculated. Based on those measurements, a scaling and a translation are applied to the MRI atlases before the registration.

\vspace{-0.3cm}
\subsubsection{Atlas-based MRI/3D US registration.}
Multimodal registration between 3DUS and MRI images is performed with a locally linear correlation metric ($LC^2$) by \cite{fuerst2014automatic} which correlates MRI intensities and gradients with US intensities.  For registration purposes, several MRI atlases of infants were combined in order to take into account anatomical variability, which included a 1yr atlas Cincinnati imaging center \cite{altaye2008infant}, a 2-5 months atlas from the McConnel Brain Imaging Center \cite{fonov2009unbiased} and 9 MRI volumes  from the ALBERTs pediatric atlas \cite{gousias2012magnetic}. 

The registration includes a rigid step with $LC^2$ and non-rigid step with $LC^2 + P$ where $P$ is a pixel weighting term.  $P$ is a term created specifically to describe lateral ventricles in US by making use of the hypoechoic area (fluid cavities) and the hyperechoic area (choroid plexus). Since only the US voxels included in the MRI ventricle label are analyzed, $P$ is only added at the non-rigid  registration step when the MRI labels are already roughly aligned to the US image:
\begin{equation}\label{Pterm}
P=\frac{C_1\sum_{i=1}^{N}\epsilon_i max(I_L-I(v_i),0)+(1-\epsilon_i)max(I(v_i)-I_H,0)+C_2}{N}
\end{equation}
where $\epsilon_i = 1$ when $ v_i$ is in the hypoechoic area and $\epsilon_i = 0$ when $v_i$ is in the hyperechoic area, $C_1, C_2$ are coefficients adjusted to the intensities and $N$ is the number of voxels in the MRI ventricle label. Moreover, $P$ is adjusted to penalize smaller labels (which statistically have higher $P$) $P_{adj}(V_k)=P(\frac{V_k}{V_M})^{\frac{1}{4}}$ where $V_k$ is the active label volume and $V_m$ the mean label volume.

The optimization of the registration process was performed using BOBYQA from \cite{powell2009bobyqa} as proposed in \cite{fuerst2014automatic}, which does not require the metric's derivatives. Registration is repeated on the 11 MRI atlases and a selection of the top ranking ($n$=4) exemplars is performed based on the resulting similarity metric. The fusion of registrations is accomplished with STAPLE \cite{warfield2004simultaneous} in order to create a probabilistic output of the labels on the 3D US images. A binary label for the lateral ventricles is then computed from the probabilistic map with all voxels having more than 80\% probability of belonging to the lateral ventricles. 
\vspace{-0.3cm}

\subsubsection{Deformable mesh model.}
Following the extraction of the binary labels based on the fusion process using STAPLE, morphological operations were applied to smooth the binary labels before it was converted to a surface mesh with a marching cubes algorithm. The mesh surface was sub-sampled to reduce computational complexity, by re-ordering the priority queue of mesh vertices and retriangulating the final mesh. Laplacian smoothing was performed on the mesh to smooth the surface prior to computing the normal vectors.

The mesh is deformed in an iterative fashion by minimizing the energy $E = E_I + \beta E_E$ where $E_I$ represents the internal energy of the system acting as a regularizer for the deformation and $E_E$ represents the external energy of the system which drives to deform the mesh. The internal energy is defined as:

\begin{equation}\label{energymesh}
E_I=\sum_{n=1}^{N_e} d (\Delta D_{1_n},\Delta D_{2_n})
\end{equation}
where $N_e$ is the number of edges, $d(.,.)$ the Manhattan distance in three dimension, with $\Delta D_{1_n}$ and $\Delta D_{2_n}$ the displacement of the first and the second vertex of edge $n$ relatively to their initial position, respectively.  

For every vertex, the term $P$ in Eq.(2) is computed for the transformed mesh as $P_{transform}$ and for the initial mesh as $P_{initial}$. The external energy $E_E$ is computed as follows: 
\begin{equation}\label{energymesh}
E_E =-\sum^{N_v}_{i=1} \left\{ \begin{array}{rcl}\sqrt{P_{transform}} \Delta D_i \gamma & \mbox{if}
& P_{initial}>=0.4 \lor \Delta D_i>0\\ \sqrt{P_{transform}} \frac{1}{|\Delta D_i|} \gamma & \mbox{if} & P_{initial}<0.4 \land  \Delta D_i<0 \\
\end{array}\right.
\end{equation}
  where $N_v$ is the number of vertices, $\Delta D_i$ is the displacement of vertex $i$ , $\gamma=1$ 
if $|\Delta D_i|<l$ ($l$ threshold set according to initial mesh) and $\gamma=\frac{1}{|\Delta D_i ^2|}$ otherwise. 

The BFGS-limited memory version optimization algorithm is used to minimize the energy equation which is well suited for optimization problems with a high number of parameters.  

\subsection{Ventricles/Brain Volume Ratio}
Once the volumes of the lateral ventricles $V_{lat.ven}$  and the total brain $V_{brain}$ are obtained, the volume ratio can be computed as follows $ratio= \frac{V_{lat.ven}}{V_{brain}}$.

\section{Results}

\subsection{Brain volume comparison between 3D US and MRI }
\textbf{Parameter selection.} Based on the comparison between the ellipsoid volume and the ground truth brain volumes in 10 MRI infant templates, optimal results were achieved when $C_f=0.95$, meaning the brain volume represents 95\% of the ellipsoid using leave-one out cross validation. 
A mean absolute difference of 2.7\% and a maximum absolute difference of 4\% was found between the estimated and ground truth brain volumes on the 10 examples used for $C_f$ determination. The 10 MRI volumes were atlases of infants all under 1 years old.

\textbf{Manual segmentation.} To first assess the agreement between modalities, brain volumes were manually extracted in 3D US by an experienced neuro-radiologist and compared to the MRI reference brain volume (mean and standard deviation of 757$^3\pm$195cm$^3$). Populations were normally distributed based on Shapiro-Wilks tests. A correlation of $r=0.988$ was found between 3D US brain volume and MRI volume. There were no statistically significant difference between both distributions based on $T$-test ($p=0.309$) and $F$-test ($p=0.477$). The mean absolute error was $3.12\pm2.65\%$.

\textbf{Automatic segmentation.} Finally, automatic brain volume assessment was performed in 3D US on the same 12 patients. Between 3D US and MRI, the correlation was $r=0.942$, with no statistically significant difference between both distributions ($T$: \textit{p}=0.541) ($F$: \textit{p}=0.273). The mean and standard deviation of the absolute errors was $7.73\pm7.52\%$ with the maximum error on a patient with abnormal brain volume due the approximation of the ellipsoid size. 
\subsection{Lateral ventricles volume comparison between 3D US and MRI}
For the comparison in ventricular volumes, manual segmentation was performed in 11 out of the 12 patients, as the US image quality was poor for a patient nearing 9 months in age. The segmentations were further validated by an experienced pediatric neuro-radiologist. Compared to the reference MRI (median of 5975mm$^3$, with a mean volume of 11084mm$^3$), there was a strong correlation in lateral ventricular volumes between 3D US and MRI ($r=0.999$), and there was no statistically significant difference between both distributions based on mean paired $T$-test and variance $F$-test ($T$: $p=0.204$) ($F$: $p=0.429$). The mean and standard deviation of the absolute differences was $5.8\pm4.92\%$. The worst individual result out of the 11 patients was due to the poor image quality linked to the infant's age ($8.5$ months) which is expected since the fontanelle opening is reduced due to bone maturation. 

Then, automatically extracted lateral ventricular volumes in 3D US were compared to the ground truth MRI volumes on the cohort of 11 patients (mean and standard deviation MRI volumes were 5309$\pm$985mm$^3$). Two images had poor image quality, and one image had ventricles dilated to almost $5\%$ of ratio and no MRI template could fit the 63 055mm$^3$ lateral ventricle volume, showing the need to add more examples to the MRI brain template. Segmentation parameters were found empirically as $C_1=0.02$,  $C_2=0.25$, $I_L=85+(I_{mean}-100)$, $I_H=115+(I_{mean}-100)$ where $I_{mean}$ is the mean intensity of the US image non zero voxels, $\alpha=0.18$ , $\beta=0.82$  and $L=2\frac{V_k}{V_M}$. For the volume comparison, a strong correlation ($r=0.848$) and no statistically significant difference were found based on $T$-test ($p=0.067$) and $F$-test ($p=0.276$) although there is a small under evaluation for the volume in 3D US (mean signed error of -6.91\%). Absolute errors have a mean and standard deviation of $9.84\%\pm4.61\%$. 
\vspace{-0.5cm}

\subsection{Segmentation of lateral ventricles in 3D US}
The segmentations were also performed on 5 additional patients with 3D US for a total of 16 infants (mean volume: 6468$\pm$320mm$^3$, max:13890mm$^3$). The accuracy was computed using expert manual segmentations as ground truth measures and the correlation with the automatically extracted volumes was $r=0.972$.  Table 1 summarizes the results with the Dice coefficient, the mean absolute distance (MAD) and the maximal absolute distance (Hausdorff).  The results demonstrate a statistically significant improvement of the proposed method to STAPLE ($p=0.0004$ for Dice coefficient and $p=0.0016$ for MAD measures), as well as to the Atlas-based approach with mesh modeling ($p=0.0059$ for Dice coefficient and $p=0.0032$ for MAD measures). Fig. 2 illustrated two examples of segmented lateral ventricles lateral with color-coded error maps representing the surface distances from the ground truth.

\vspace{-0.5cm}

\begin{table}[h]
\begin{center}
\caption{\small  Comparison in accuracy of the lateral ventricular segmentation methods from 3D US, based on Dice coefficients, mean absolute distance and Hausdorff distance.}
\small
\begin{tabular}{lcccc}
\hline
 {{Methods}}  &   {{DICE (\%) }} &  {{MAD(mm)} }  &  {{Hausdorff(mm)} }\\
\hline
Atlas-based with $LC^2$ [5]  & \rmfamily $57.4\pm 7.8$  &  \rmfamily$1.33\pm0.44$  &  \rmfamily  \rmfamily$8.55\pm3.42$  \\
Atlas-based with area weights [5]  & \rmfamily $60.4\pm 7.5$  &  \rmfamily$1.14\pm0.30$  &  \rmfamily  \rmfamily$7.52\pm2.81$  \\
Atlas-based [5] + Mesh  & \rmfamily   $65.1\pm 4.1$  & \rmfamily$1.08\pm 0.33$ &   \rmfamily$8.46\pm2.98$ \\
Majority Voting (MV)  & \rmfamily $65.0\pm 4.0$  &  \rmfamily$1.01\pm0.30$  &  \rmfamily  \rmfamily$7.59\pm3.40$  \\
STAPLE [10] & \rmfamily $65.5\pm 3.8$  &  \rmfamily$1.08\pm0.24$ &  \rmfamily  \rmfamily$7.27\pm3.19$  \\
\textbf{Proposed method} & \rmfamily   $70.8\pm 3.6$ & \rmfamily$0.88\pm 0.20$ &   \rmfamily$6.84\pm3.15$ \\
\hline
\end{tabular}
\end{center}
\end{table}
\vspace{-1cm}

\vspace{-0.7cm}
\begin{figure}[h]
\begin{center}
\includegraphics [scale=0.35]{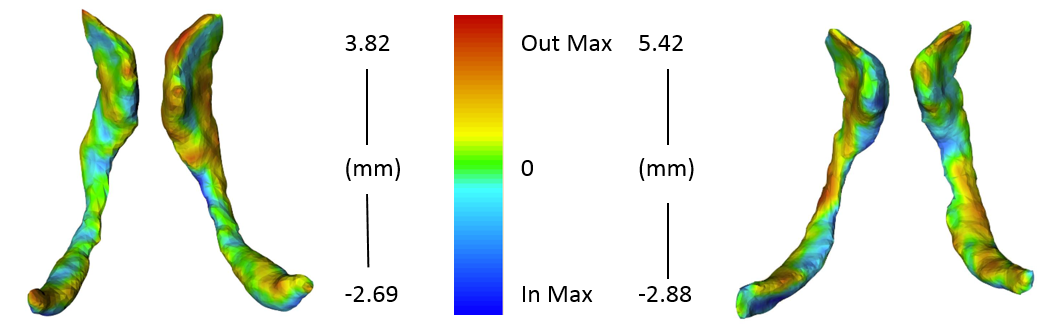}
\caption{\small  Two examples of segmented lateral ventricles with error maps from 3D US.}
\end{center}  
\end{figure}

\vspace{-1cm}
\subsection{Ventricular-total brain volume ratio in 3D US}
For the 12 infants of the prospective cohort, the ground-truth ventricular ratios were computed from the MRI segmentations of the lateral ventricles and estimation of the brain volume. For the ratios from 3D US, the median difference was $0.00795$, with a mean difference of $0.0125\pm0.0144$. 
In terms of concordance between MRI and 3D US, a correlation of $r= 0.998$ demonstrate the strong agreement, and there was no statistically significant difference based on paired $T$-test ($p=0.672$) and $F$-test ($p=0.437$). The absolute errors yielded a mean and standard deviation of $6.05\pm4.88\%$.
\vspace{-0.5cm}

\section{Conclusion}

In this paper, we presented an automatic method to extract lateral ventricles as well as total brain volumes from 3D ultrasound in infant brains. This allows for an automatic assessment of the lateral ventricles dilatation with respect to total brain volumes. Compared to MRI references, the volumes yielded a high correlation and indicate no statistically significant difference between both modalities. In addition, volume ratios can be obtained with a mean ratio slightly below 0.01, which is concordant with literature. 
Our main contribution is the quantification of the ventricular-brain ratio in 3D US which enables a true assessment of ventricular dilatation.  Future work would include adding more MRI templates and an extensive validation with additional subjects, both with higher variability in ventricular volumes, and investigate the use of convolutional neural networks.

\bibliographystyle{splncs}
\bibliography{biblio}

\end{document}